\pdfoutput=1

\documentclass[11pt]{article}

\usepackage[]{acl}

\usepackage{times}
\usepackage{latexsym}
\usepackage[T1]{fontenc}
\usepackage[utf8]{inputenc}
\usepackage{microtype}
\usepackage{graphicx}

\title{Semantic Search as Extractive Paraphrase Span Detection}

\author{Jenna Kanerva$^{1}$, Hanna Kitti$^{1}$, Li-Hsin Chang$^{1}$, Teemu Vahtola$^{2}$, \\
        \textbf{Mathias Creutz$^{2}$, and Filip Ginter$^{1}$} \\
  $^{1}$ TurkuNLP, Department of Computing, University of Turku, Finland\\
  $^{2}$ Department of Digital Humanities, Faculty of Arts, University of Helsinki, Finland}

\begin{document}
\maketitle
\begin{abstract}
In this paper, we approach the problem of semantic search by framing the search task as paraphrase span detection, i.e.\ given a segment of text as a query phrase, the task is to identify its paraphrase in a given document, the same modelling setup as typically used in extractive question answering. On the Turku Paraphrase Corpus of 100,000 manually extracted Finnish paraphrase pairs including their original document context, we find that our paraphrase span detection model outperforms two strong retrieval baselines (lexical similarity and BERT sentence embeddings) by 31.9pp and 22.4pp respectively in terms of exact match, and by 22.3pp and 12.9pp in terms of token-level F-score. This demonstrates a strong advantage of modelling the task in terms of span retrieval, rather than sentence similarity. Additionally, we introduce a method for creating artificial paraphrase data through back-translation, suitable for languages where manually annotated paraphrase resources for training the span detection model are not available.
\end{abstract}

\section{Introduction}

With the existence of large, pre-trained language models, such as BERT~\citep{devlin-etal-2019-bert}, GPT~\citep{radford2019language}, or T5~\citep{raffel2020t5}, numerous NLP task requiring deep language understanding have recently gained promising results. For example, in natural language inference and question answering such models have helped to substantially narrow down the gap between human and model performance (see e.g. \citet{sun2021ernie} or \citet{raffel2020t5}). One task clearly requiring deep language understanding is semantic search, where the objective is to retrieve from a document those passages that match the search query in their meaning, rather than in their surface forms only.

Semantic search can also be seen as a form of paraphrase detection, identifying statements equivalent in meaning but differing on the surface level. While the traditional term-based search techniques are to a large extent limited to returning results based on surface form matching, the hope in semantic search is to rather understand the key meaning of the search phrase and return the relevant knowledge. For example, when querying using the phrase \emph{What are the dimensions of Volkswagen Transporter} also documents mentioning the paraphrased versions \emph{VW Transporter: size} or \emph{the length, width and height of VW Transporter} should be considered relevant. 

Recently, a large-scale corpus of Finnish paraphrases, the Turku Paraphrase Corpus~\citep{kanerva2021paraphrase}, became available. The paraphrase pairs in the corpus are manually extracted from pairs of related documents, forming annotated examples where the document context of both members of the paraphrase pair is known. This very property of the dataset is to the best of our knowledge unique to this corpus and in turn allows us to take a novel approach to semantic search, by casting it as paraphrase span detection: Given a segment of text as a query, the task of the model is to identify its paraphrase from the given document. Span detection is typically used in extractive question answering. The primary advantage of using span detection, as opposed to the conventional approach of classifying sentence pairs or computing their pairwise similarity, is the ability to easily extract any part of the target document, not just predefined units such as lines or sentences.

We evaluate the span detection model trained on the Finnish paraphrase data and compare it to two sentence-level retrieval baselines. Additionally, we introduce a straightforward method of generating artificial paraphrase data through back-translation, allowing training also for languages where manually annotated paraphrases-in-context data is not available. Finally, we carry out an extensive error analysis to understand the prediction capabilities of the span detection model.


\section{Related Work}


\paragraph{Paraphrase} In NLP, different paraphrase related tasks include detecting, retrieving or generating paraphrased versions of a given text span. Numerous paraphrase corpora, e.g. Quora Question Pairs\footnote{\url{data.quora.com/First-Quora-Dataset-\\Release-Question-Pairs}}, Microsoft Research Paraphrase Corpus \citep{dolan2005MSRP}, and PARADE \citep{he-etal-2020-parade}, have been released for these purposes, each including labeled sentence-like text pairs supporting mainly paraphrase classification. \textbf{Paraphrase retrieval} is typically approached using large monolingual corpora and performing one-to-many (find a paraphrase for the given text span) or many-to-many (find all paraphrase pairs from the text collection) sentence similarity comparison between calculated sentence-level embeddings (see e.g. \citet{info11050241}), potentially including document level heuristics in order to restrict the search space into comparable documents. To our knowledge, besides the Turku Paraphrase Corpus, there is no other paraphrase corpora available where the original document context would be available for the paraphrase pairs, and thus directly supporting the paraphrase span detection task. 


\paragraph{Question Answering} In extractive question answering, the system is given a question posed in natural language together with a background document, and the task is to extract the span of the correct answer from the document. Span detection is a common approach for extractive question answering, naturally supporting extracting an answer segment of any length. There are both monolingual and cross-lingual question answering (QA) datasets available. SQuAD \citep{rajpurkar2016squad} is an English QA dataset including approx.\ 100,000 examples where the context document has an answer for the given question. In its second release (SQuAD v2), also unanswerable questions are included \citep{rajpurkar2018squad2}. Some multilingual QA corpora include e.g.\ XQuAD \citep{artetxe-etal-2020-cross}, TyDiQA \citep{clark-etal-2020-tydi}, and MKQA \citep{mkqa}, the latter two including also Finnish QA examples. Even though the task setup used in this work resembles the QA task, the objective is different. While in QA the system is expected to return an answer for the question, in paraphrase retrieval it returns a semantically equivalent segment from the background document.


\paragraph{Semantic Textual Similarity} In the semantic textual similarity (STS) task, each sentence pair is annotated with a similarity score typically ranging from 0 to 5, where lower scores mean unrelated or related sentences, while higher scores are for partially or fully equivalent sentences, with the highest score typically indicating the sentences being completely equivalent in meaning. The annotations in STS and paraphrasing tasks are highly related \citep{gold-etal-2019-annotating} but not necessarily completely interchangeable between different datasets as the definition of a paraphrase or relatedness may not be fully equivalent. Similar to paraphrase datasets, most of the STS datasets include pairs of approximately sentence-long text snippets together with the annotated degree of similarity, therefore supporting the setting of a sentence-pair classification task without contextual information (see e.g. \citet{agirre-etal-2016-semeval,cer-etal-2017-semeval}). However, a recent dataset of \citet{sido2021czech} includes similarity annotations of Czech sentence pairs in document context, thus to our knowledge being the first STS dataset which could directly be applied to span classification modelling.



\section{Data}

\begin{figure*}
\centering
\includegraphics[width=0.9\textwidth]{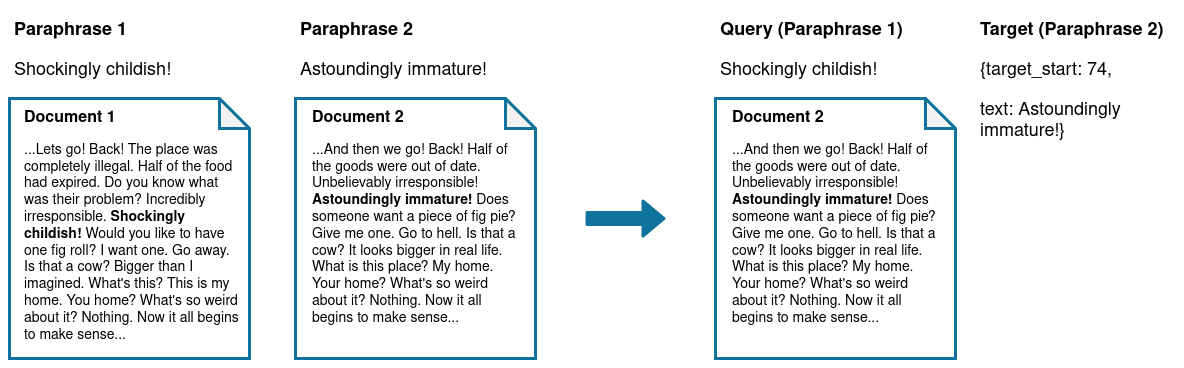}
\caption{On the left side is an illustration of one paraphrase pair from the Turku Paraphrase Corpus, and on the right side is the same paraphrase pair turned into the span detection framework as used in this work.}
\label{fig:task-setup}
\end{figure*}

The Turku Paraphrase Corpus\footnote{Newest data release available at: \url{https://github.com/TurkuNLP/Turku-paraphrase-corpus}} consists of paraphrase pairs manually extracted from pairs of related documents with high probability for naturally occurring paraphrases. As mentioned previously, the position in the respective source documents is preserved. Most of the pairs are obtained from independent subtitle versions of the same movie or TV episode. Subtitles thus constitute the primary domain of the data, while a small portion is extracted from other domains, including news articles, discussion forum messages as well as university exercises and essays. Furthermore, each paraphrase pair is manually categorized in a scheme distinguishing paraphrases primarily by the degree of their context independence. In Figure~\ref{fig:task-setup} we illustrate one paraphrase pair from the original corpus, as well as its transformation into the span detection setting used in this work.

The corpus has two categories of examples of interest for this study: 86,986 positive examples of naturally occurring paraphrases in their respective document contexts and 1,308 negative examples of pairs in their document contexts that are semantically similar but not mutual paraphrases. These constitute 84\% of the corpus. The remaining 16\% are unsuitable for this study as they either do not have document contexts for various reasons, or are manually edited and therefore no longer naturally fitting their contexts.

As the paraphrase pairs are not directional in the same manner as for example question-answer examples are, and the two paraphrases are always extracted from two distinct context documents, each pair produces two distinct examples in the span detection task, resulting in a total of 173,972 distinct positive and 2,616 distinct negative examples. The data statistics are summarized in Table~\ref{tab:data-size} in terms of train, development and test sets, following the dataset split provided in the original corpus.

We pursue two different task setups: 1) Retrievable paraphrases formed from the positive examples, where for all examples a valid paraphrase is guaranteed to exist in the context. The setup is similar to SQuAD v1 in question answering. 2) Including the 2,616 negative examples as irretrievable paraphrases, requiring the model not only to find a valid paraphrase, but also being able to determine when there is not a valid paraphrase present in the context. The setup is similar to SQuAD v2 in question answering.

\begin{table}[]
    \centering
    \begin{tabular}{l|c|c}
                & Setup 1        & Setup 2      \\
       Section  & Examples       & Examples   \\\hline
       Train    & 138,706        & 140,848      \\
       Devel    & 17,702         & 17,930     \\
       Test     & 17,564         & 17,810      \\\hline
       Total    & 173,972        & 176,588    \\
    \end{tabular}
    \caption{Dataset sizes after converting the original paraphrase data into the span detection framework, where Setup 1 includes only retrievable examples, while Setup 2 includes both retrievable and irretrievable examples.}
    \label{tab:data-size}
\end{table}

\section{Experiments}

\subsection{Paraphrase-SD model}

The span detection model described in this section and referred to as \emph{Paraphrase-SD} throughout the paper is based directly on the implementation of the question answering task with a BERT encoder in the well-known HuggingFace library\footnote{\url{https://github.com/huggingface/transformers}} \citep{wolf-etal-2020-transformers}. Given a query phrase and a document, separated by the [SEP] token, the model detects the span in the document which paraphrases the query as follows: Each subword encoded by the BERT model is classified by two classification layers, one for predicting the start position of the span and one for predicting the end position. Both output layers are binary and applied independently, thus individually predicting how likely each subword is opening and/or closing the target span. The output of the model is then the span which maximizes the sum of the logits for its opening and closing subword. In order to be a valid span, the end position must be higher or equal to the start position, and the start position must point to the context region of the input (sequence after the [SEP] token), not the query phrase region. In Setup 2 that includes also irretrievable examples, the model must also be capable of empty predictions. For these, the model is trained to predict the [CLS] token as both start and end position of the span, thus in practise returning an empty span (null prediction).

Many of the documents are longer than the maximum sequence length of the BERT model. We slice the documents (with overlap of 128 tokens) into segments, which form independent examples. Each of these examples thus consists of the query phrase which is never sliced nor truncated, and a slice of the document. Predictions for these examples are subsequently merged into a single prediction for the whole document, as follows: In Setup 1 including only retrievable paraphrases, the span with the highest aggregated score out of all possible spans over all slices of the document is chosen. However, this necessary slicing interferes with Setup 2. When there are multiple document slices, the model is likely to give a highly confident null prediction for all slices not including the target span. When aggregating the scores across all slices of the document, these confident null predictions would dominate the output. Noting that all document slices give some probability for the null prediction, the final null prediction score can be obtained by taking the minimum value (least confident null prediction) across all document slices, approximating the null prediction value obtained for the full document at once. That score is then compared against the most confident span predictions selecting the span with the highest overall value as the final prediction.


We use the HuggingFace transformers library question answering model implementation, with the Finnish \emph{FinBERT}~\citep{virtanen2019multilingual} language model as the encoder. The weights of the pre-trained language model are fine-tuned together with the two task specific classification layers during training. We performed a grid search separately for Setup 1 and Setup 2 in order to find optimal hyperparameters. Trialed hyperparameters were batch sizes 8, 16 and 32, learning rates 5e-5, 3e-5 and 2e-5 and epochs 2 and 3 on development section of the data. For all experiments with Setup 1 we use batch size 32, learning rate 3e-5 and the model is trained for two epochs. Respectively, for all the experiments with Setup 2 the hyperparameters are: batch size 16, learning rate 2e-5 and two epochs. The source code is available at \url{https://github.com/TurkuNLP/paraphrase-span-detection}.

\subsection{Baselines}

We compare the Paraphrase-SD model with two baselines. The first is a straightforward tf-idf baseline, where for each paraphrase in the evaluation data, the most similar sentence in the target document is retrieved based on the cosine similarity of tf-idf weighted vectors. We tested word-level features as well as character n-grams created inside word boundaries and maximum number of features set to 300,000. N-gram lengths in the range [2,6] were systematically tested and the best results were gained with the union of character n-grams of lengths 2, 3 and 4. The n-gram vocabulary was induced on the training data only.


Our second baseline is based on FinBERT model embeddings. Similar to the first baseline, for each query, the most similar sentence in the document is retrieved, with the embedding for each sentence calculated as the average of token embeddings obtained from the last hidden layer of the FinBERT model without any fine-tuning. Again, cosine similarity of the embeddings is used to calculate the similarity measures.

One notable advantage of the Paraphrase-SD model is its ability to return any text segment from the background document. Both baselines, on the other hand, are limited to sentence level predictions, in order to avoid embedding all possible document segments of any length, which would be highly impractical. However, as the paraphrases in the Turku Paraphrase Corpus are not strictly limited to sentence boundaries, with about 25\% being longer or shorter than a sentence, the baseline approaches incur a loss. To assess its magnitude, we will report also the oracle performance, corresponding to returning the one sentence from the document that is most overlapping with the true target span.

\subsection{Paraphrase-SD through back-translation}

Up to this point, we relied on the fact that the Turku Paraphrase Corpus enables our approach by containing paraphrases in their context. Such a dataset is, to the best of our knowledge, currently available only for Finnish. In this section, we explore a straightforward heuristic approach based on a form of back-translation~\citep{sennrich-etal-2016-improving}, allowing the application of the Paraphrase-SD model also in absence of such a manually annotated corpus.

We take an approximate of 60K Finnish subtitle files from the same subtitle domain as in the original Turku Paraphrase Corpus that were not used in the original data, and split them into shorter text segments yielding 260K documents. Additionally, we have acquired approximately 200K Finnish documents from the Reddit discussion forum to accompany the data.
For each of these documents, we randomly sample one sentence from any position in the document to act as a target sentence whose span is to be retrieved from the document in the span detection paraphrase retrieval task. We remove examples consisting of target sentences longer than 100 word tokens to reduce unnecessarily noisy examples. Finally, we use back-translation to generate an assumed paraphrase for each sampled retrievable target sentence. We translate the original Finnish target sentences into English, and back into Finnish using pre-trained translation models from the OPUS-MT project~\citep{TiedemannThottingal:EAMT2020}. We decode the translations using beam search with a beam size of 6 and a length normalization term of 0.6 in both directions. We collect the most probable back-translated sentence for each source sentence to act as a paraphrase of the original sentence. The back-translated sentence is always used as the query phrase, while the original sentence in its context acts as the retrievable target span. 

The back-translated data is used to train the span detection model using the same hyperparameters as with the original model. The back-translated data was randomly sampled to the same size as the original training data, however before sampling we removed examples where the back-translation produced an identical sentence compared to the target span, an empty sentence, or a sentence longer than 380 subwords\footnote{Sentences longer than 380 subwords were filtered out due to preserving enough space for the document in the model's input.}. Since the back-translated data contains only retrievable paraphrases, it is used in the Setup 1 experiments only. The final size of the training data for the back-translation baseline is 138,706 examples.


\section{Results}

\begin{table*}[]
    \centering
    \begin{tabular}{lcc|cc}
    & \multicolumn{2}{c}{Setup 1}   & \multicolumn{2}{c}{Setup 2} \\
      Model                                   & EM & F-score  & EM & F-score \\\hline
      Sentence-level baselines                & & & & \\
      \hspace{3mm}TF-IDF                      & 56.84 & 72.02 & 56.06 & 71.03  \\
      \hspace{3mm}BERT                        & 66.32 & 81.44 & 65.40 & 80.31  \\
      \hspace{3mm}Oracle                      & 76.74 & 93.85 & 75.70 & 92.57 \\\hline
      Paraphrase-SD                           & & & &\\
      \hspace{3mm}Back-translation             & 71.32 & 85.07 & ---   & --- \\
      \hspace{3mm}Main model                   & \textbf{88.73} & \textbf{94.31} & \textbf{84.37} & \textbf{89.52} \\
      Data augmentation                 & & & &\\
      \hspace{3mm}+ Artificial irretrievables                 & ---   & ---   & 82.35 & 87.11 \\
      \hspace{3mm}+ Back-translation (random)                 & 88.67 & \textbf{94.48} & --- & --- \\
      \hspace{3mm}+ Back-translation (tf-idf 0.35-0.66)       & 88.61 & 94.28 & --- & --- \\
      \hspace{3mm}+ Back-translation (tf-idf most dissimilar) & 88.38 & 94.24 & --- & --- \\\hline

    \end{tabular}
    \caption{The main results for the Setup 1 including only retrievable paraphrases as well as for Setup 2 where both retrievable and irretrievable paraphrases are used. Results are reported in terms of exact match (EM) and token-level F-score on test section.}
    \label{tab:main-results}
\end{table*}

The main results are summarized in Table~\ref{tab:main-results} using two evaluation metrics: \emph{exact match (EM)}, which measures the percentage of predictions that match the gold segment exactly, and \emph{F-score}, which measures the average token-level overlap between the prediction and the gold segment, when segments are treated as bag-of-tokens and punctuation characters are disregarded. All results are reported using the test section of the Turku Paraphrase Corpus (over 17,000 examples for both setups, see Table~\ref{tab:data-size}).

Our main model, Paraphrase-SD trained with the Turku Paraphrase Corpus, outperforms all baselines in both setups with a clear margin, receiving EM 88.73 and F-score 94.31 for Setup 1 (only retrievable paraphrases), and EM 84.37 and 89.52 F-score for Setup 2 (retrievable and irretrievable paraphrases). The second best performing model, Paraphrase-SD model trained with back-translation data, sees about -17.4pp decrease in EM compared to the main model in Setup 1. Both BERT and tf-idf baselines fall behind the back-translation, having -22.4pp and -31.9pp decrease respectively in EM compared to the main model. The results are similar in terms of F-score, the back-translation and the two baselines being -9.2pp, -12.9pp, and -22.3pp behind the main model. The sentence-level retrieval significantly harms the theoretical upper bound (oracle) of the baselines in terms of EM, and to a much lesser degree in terms of F-score, clearly demonstrating the intrinsic disadvantage of limiting retrieval to such pre-defined units. Nevertheless, both BERT and tf-idf baselines fall notably behind the oracle performance in both metrics, showing that the sentence-level retrieval is not the main limiting factor in the baseline performance. By comparing the Setup 1 and Setup 2 results for the main model, we can see that including the irretrievable cases in the training data and asking the model to recognize when the correct paraphrase does not exist in the document decreases the performance of the Paraphrase-SD model. Naturally, the behavior is expected due to introducing a more difficult task setup.

\subsection{Data augmentation experiments}
\label{sec:data-augmentation}

The Turku Paraphrase Corpus contains only a small fraction of non-paraphrase pairs, corresponding to irretrievable examples in our Setup 2. Therefore, we first experiment with increasing the small proportion of irretrievable examples in the original training data by automatically creating artificial irretrievable training examples for the Setup 2 model. These are created from retrievable examples by simply removing the target span from the document. Each retrievable training example is thus introduced twice in the training data, once as a retrievable example and once as an artificial irretrievable example, resulting in total of 279,554 training instances with approximately 50/50 label distribution. Nevertheless, we find that the artificial irretrievables did not improve the model performance, being approximately -2pp worse than the original model on both metrics, mostly resulting from a noticed increase in false null prediction rate. This is not a particularly surprising finding, given that the evaluation data is not changed, causing a distribution mismatch between training and evaluation data. Since the removed target span no doubt leaves an unnatural artefact in the document, in the place where the target used to be, which the model can learn to recognize, the results on such artificially modified evaluation data would not have been reliable. The fact that the decrease in performance is quite limited, even though the training data distribution is substantially altered shows a surprising resilience of the model.

Next, we carry out preliminary data augmentation experiments, where our primary training data is enhanced with the artificially created back-translation examples. We train three additional models with mixtures of original and back-translation training data using different sampling strategies, each including exactly 138,706 back-translation examples matching the size of the original training set, and thus doubling the training data compared to the original Paraphrase-SD model. The first model uses a random sample of the back-translation data, the second model strives to include ``interesting'' examples with low lexical overlap obtained by sampling the most dissimilar query--target back-translation pairs in terms of tf-idf similarity, and the third model balances between too similar (trivial examples) and too dissimilar (likely including translation errors) by sampling mid-range examples using tf-idf similarities between 0.35--0.66. The tf-idf similarities are calculated using the same parameters as for the tf-idf sentence retrieval baseline.

The results (Table~\ref{tab:main-results}) show that even with pure back-translation data, the model exceeds the performance of the BERT and tf-idf baselines, showing back-translation as a viable option for training a span-detection-based retrieval model. However, when combined with the original training data from the manually annotated paraphrase corpus, the back-translation data did not improve the overall results. The best result was obtained with the randomly sampled back-translation data, exceeding the performance of the main model by a mere +0.2pp. Our experiments of selecting more interesting back-translation examples did not yield positive results over the random selection. To maintain the focus and scope of this study, we do not proceed examining the multitude of possible strategies of sampling and incorporating the back-translation data. Nevertheless, given the positive outcome when compared to the tf-idf and BERT baselines, more detailed experiments are clearly justified as a future work.

\section{Error Analysis}

We perform several analyses on the development data in order to better understand the capabilities of our main model and the reasons behind incorrect predictions. Firstly, we automatically categorize the incorrect predictions into several subgroups and inspect the different error groups. Secondly, we calculate the prediction accuracy against the estimated paraphrase complexity in order to investigate for example whether certain paraphrase categories in the data include more incorrect predictions than others. These experiments are carried out using the development section of the Turku Paraphrase Corpus. Finally, we test the out-of-domain generalization of the model by dividing the paraphrase data into two distinct domains. As the out-of-domain analysis does not require any manual inspection, it is carried out using the test section of the corpus, the numbers then being comparable with the main experiments.

\subsection{Error categorization}

The incorrect predictions as determined in terms of the exact match measure are categorized into several subgroups: 1) the model gave empty prediction even if a valid target exists in the document (false null prediction for Setup 2), 2) the model predicted a span matching one of the negative paraphrase pair examples in the corpus, 3) the model predicted a span partially overlapping with the target, further divided into three subgroups: 3.1) the prediction is a substring of the gold segment, 3.2) the gold segment is a substring of the prediction, 3.3) other partial overlap in predicted and gold segments, and 4) other, including cases where the model predicts a segment not overlapping with the gold annotation. The distribution of mispredictions categorized into subgroups is given in Table~\ref{tab:error_groups}.


\begin{table}[]
    \centering
    \begin{tabular}{lcc}
      Misprediction type                         & Setup 1   & Setup 2 \\\hline
      (1) Null prediction                        & ---       & 36.50\% \\
      (2) Pred not-paraphrase                    & ---       & 7.17\%  \\
      (3) Partially correct pred                 & 58.63\%   & 38.67\% \\
      \hspace{2mm}(3.1) pred substr. of gold     & 35.52\%   & 23.86\% \\
      \hspace{2mm}(3.2) gold substr. of pred     & 22.29\%   & 14.16\% \\
      \hspace{2mm}(3.3) other partial overlap    & 0.81\%    & 0.65\%  \\
      (4) Other                                  & 41.37\%   & 17.67\% \\\hline
   
    \end{tabular}
    \caption{The error categories of incorrect predictions on the development data.}
    \label{tab:error_groups}
\end{table}

While the errors categorized as subgroup (1) and subgroup (2) can be seen as clear mispredictions, where the model is not able to identify a paraphrase even if one is guaranteed to exist in the document, or identifies the segment annotated as not-a-paraphrase in the original data, the errors belonging to the subgroup (3) contain cases of partially correct predictions where the model is able to identify approximately the correct area from the document, however, the predicted start and end positions slightly differ from the gold segment. From the finer subcategorization it can be seen that when the model makes a partially correct prediction, it is more likely to exclude some part of the gold segment rather than include an additional part. The number of other partial overlaps is negligible.


On the other hand, mispredictions in the subgroup (4) include cases which require further manual evaluation in order to determine their correctness. In these cases the model suggests a paraphrase candidate the annotators have not extracted during the corpus construction. These predictions cannot be directly determined to be incorrect, as it is possible that the document includes another occurrence of a paraphrase of the query, which the model then extracts. For many common generic phrases, the probability of more than one correct paraphrase existing in the document is not negligible. Therefore, the evaluation can be considered to give only the lower bound slightly underestimating the actual performance. Therefore, we perform a further manual evaluation for the category (4) predictions.

We sample 200 incorrect predictions from the subcategory (4) for the Setup 1 model, and manually annotate for each example whether the predicted span is a valid paraphrase for the query phrase. We find that full 36\% of these are in fact valid paraphrases of the query although not being the gold target segment, mostly due to short repeating lines and generic phrases in the movie subtitle section of the corpus, or repeating material between the title, the lead paragraph and the article body in the news article section of the corpus.





\subsection{Paraphrase complexity}
\label{sec:estimated-difficulty}

The paraphrase corpus classifies each paraphrase into one of several classes: \emph{Context dependent} are mutual paraphrases in their present context but not necessarily in other contexts, \emph{context independent} are perfect mutual paraphrases in all reasonably imaginable contexts. In between these two categories, there are near-perfect context independent paraphrases up to one or more qualifying flags: \emph{style} for tone or register difference, \emph{minor difference} marking easily traceable grammatical differences such as person and number, and \emph{subsumption} marking there is a degree of directionality in the relation, with for instance one paraphrase mentioning \emph{a woman} while the other \emph{a person}. The reader is referred to the annotation guidelines \citep{kanerva2021annotation} for more detailed descriptions and examples of these classes. 

This classification allows us to inspect the model performance w.r.t.\ the type of the paraphrase, and its degree of context dependence. As seen in Table~\ref{tab:error_by_label}, indeed the Setup 1 model performance clearly correlates with the ``degree of universality'' of the paraphrases, with 13pp difference between perfect universal paraphrases and context dependent paraphrases.

Given the hypothesis of simpler paraphrases resulting in more confident predictions, Table~\ref{tab:error_by_category} shows the prediction accuracy across automatically classified "trivial" paraphrase categories following \citet{chang2021quantitative}. Here, paraphrases are considered trivial if all their differences can be accounted for with simple, automatically recognizable transformations. These categories include phrases which share the same lemmas thus differing only in word order or inflections, have the same lemmas in terms of content-bearing words only, or if their only differences can be accounted for with a synonym list, or a combination of these. Results show that non-trivial paraphrases, which account for most of the data, more often lead to incorrect predictions compared with trivial paraphrases. However, given the small frequency of trivial paraphrases in the data, the results may suffer from sampling bias.

\begin{table}[]
    \centering
    \begin{tabular}{lrr}
      Paraphrase type                 & Acc  & Support    \\\hline
      Context independent             & 95.0   & 3,898 \\
      \hspace{3mm}with minor diff.       & 92.5   & 890  \\
      \hspace{3mm}with style diff.          & 90.2  & 902  \\
      \hspace{3mm}with subsumption          & 89.5  & 8,372 \\
      Context dependent                & 82.0 &  4,632 \\\hline
      Overall                            & 90.2  & 17,702 \\\hline
   
    \end{tabular}
    \caption{Prediction accuracy in Setup 1 for the different paraphrase types annotated in the corpus. }
    \label{tab:error_by_label}
\end{table}

\begin{table}[]
    \centering
    \begin{tabular}{lrr}
      Category                         & Acc & Support   \\\hline
      Trivial                           & 96 & 516    \\
      \hspace{3mm}same lemmas           & 91 & 32     \\
      \hspace{3mm}same content word lemmas          & 97 & 30     \\
      \hspace{3mm}synonym replacement   & 98 & 322    \\
      \hspace{3mm}content word lemmas & & \\
      \hspace{3mm}with synonym replacement  & 96 & 132    \\
      Non-trivial                       & 90 & 17,186 \\\hline
      Overall                           & 90 & 17,702 \\\hline
   
    \end{tabular}
    \caption{Prediction accuracy in Setup 1 on several categories of trivial paraphrases. }
    \label{tab:error_by_category}
\end{table}

\subsection{Out-of-domain experiments}

The majority of the Turku Paraphrase Corpus data is collected from movie and TV episode subtitling data, only 14\% of the paraphrase pairs used in this work being from other text domains (e.g news or discussion forum). To find out how the imbalance of the training data domains affects the prediction ability we evaluated the main model (Setup 1) performance separately on evaluation data divided by the domains into two parts, subtitle and non-subtitle. As expected, the main model performs better on subtitle data, giving exact match scores of 91.72 for subtitle and 66.06 for non-subtitle. For out-of-domain generalization experiments, we also trained a model using only subtitling data from the training set, and evaluated in the same two domains in order to compare the effect of the small amount of in-domain data in the main model when evaluated on non-subtitle domain. Compared to the main model the performance of this model is only slightly decreased when evaluated on in-domain subtitle data (-0.1pp EM) likely due to the small decrease in the size of training dataset, however, the decrease is notable on the out-of-domain data (-4.9pp EM).

\section{Conclusions}

In this paper, we have taken a novel approach to semantic search by casting it as extractive span detection of paraphrases, where given a query phrase, the task is to identify its paraphrased target span from a given document. The primary advantage of such a model is its ability to retrieve a segment of any length, not just a predefined units such as sentences, as is the case with the standard retrieval methods utilizing for example embedding similarities. Our span detection model trained on the manually annotated Turku Paraphrase Corpus clearly outperformed the two retrieval baselines relying on lexical similarity or BERT sentence embeddings by 31.9pp and 22.4pp respectively in terms of exact match, demonstrating a clear advantage of the new modelling approach.

Additionally, we have introduced a method for creating artificial paraphrase data through back-translation, suitable for languages where similar paraphrase data including document context is not available. While not achieving the performance of the model trained on the manual paraphrase data, the back-translation model clearly outperforms the sentence embedding baselines.

\section*{Acknowledgements}

The research was supported by the Academy of Finland and European Language Grid. Computational resources were provided by \emph{CSC --- the Finnish IT Center for Science}.

\bibliography{custom}
\bibliographystyle{acl_natbib}


\end{document}